\documentclass{article}

\usepackage{microtype}
\usepackage{booktabs}
\usepackage{graphicx}
\usepackage{xurl}
\usepackage[hyperfootnotes=false]{hyperref}
\usepackage[preprint]{icml2027}
\usepackage{amsmath,amssymb,mathtools}
\hypersetup{pdfsubject={Preprint}}

\DeclareMathOperator{\EditOps}{EditOps}
\DeclareMathOperator{\Clip}{Clip}
\icmltitlerunning{DFPO for AI Coding Agents}
\input{dfpo-bibliography.tex}

\begin{document}

\twocolumn[
  \icmltitle{Direct Feedback Policy Optimization for AI Coding Agents (DFPO)}
  \begin{icmlauthorlist}
    \icmlauthor{Fei Ding}{alibaba}
  \end{icmlauthorlist}
  \icmlaffiliation{alibaba}{Alibaba Group}
  \icmlcorrespondingauthor{Fei Ding}{\mbox{dignfei@gmail.com}}
  \icmlkeywords{AI Coding, Agent Reinforcement Learning, Credit Assignment}
  \vskip 0.3in
]
\printAffiliationsAndNotice{}

\begin{abstract}
AI coding agents generate, modify, and validate code through multi-turn interactions.
Existing reinforcement learning methods often rely on sparse terminal rewards, making it difficult to assess how intermediate decisions affect task outcomes, and group-relative methods also require repeated sampling for each task.
We propose Direct Feedback Policy Optimization (DFPO), a single-trajectory reinforcement learning algorithm that trains directly from a single trajectory.
DFPO achieves higher task resolution rates than GSPO under matched base models and execution harnesses.
\end{abstract}

\section{Introduction}
AI coding agents must interpret requirements, implement features, modify code, and validate their work in a real workspace. Executable tests provide acceptance signals for these tasks. Repository repair benchmarks, for example, require a final patch to fix the target issue while preserving existing behavior \citep{arx2310_06770,arx2505_20411,sweverified}.

A complex repository repair task can take about an hour before final validation reveals the outcome. In the terminal-reward setting, the learner receives a single task-level reward for the entire episode, issued only at termination. An execution trajectory often contains useful implementations, unproductive attempts, and edits that introduce regressions. This sparse and delayed reward makes it difficult to tell which intermediate decisions contributed to success and which led to failure. Assigning the same terminal reward to all generated tokens updates these behaviors in the same direction.

\begin{figure*}[t]
  \centering
  \includegraphics[width=\linewidth]{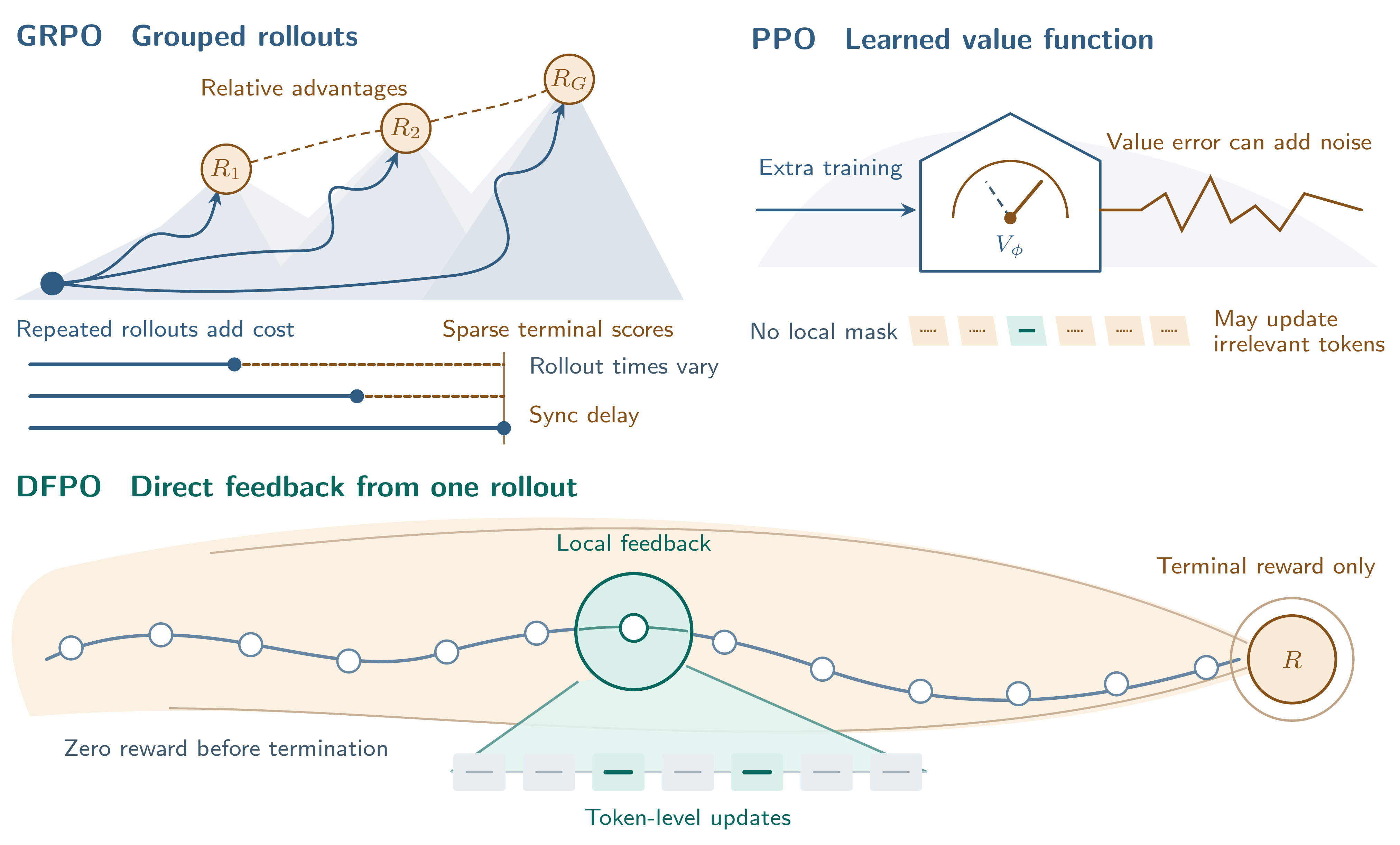}
  \caption{Sources of credit assignment. The three mountain paths show independent rollouts for the same task. The summit markers are sparse terminal scores, from which GRPO estimates group-relative advantages \citep{arx2402_03300}. Solid lines below the paths show unequal rollout durations. Dashed lines show how long earlier rollouts wait for the slowest under synchronous grouping \citep{arx2606_02218}. The PPO gauge and waveform illustrate how value estimation errors can introduce noise into advantage estimates \citep{arx1707_06347,arx1506_02438}. Orange token blocks mark irrelevant positions that may be updated without a local mask. The DFPO wave shows one task trajectory. The environment reward is zero before termination, and the task receives score $R$ at the endpoint. The teal magnifier shows how local feedback focuses updates on a few tokens.}
  \label{fig:dfpo-overview}
\end{figure*}

A failed test provides feedback that can be reused. Given the original output at the point of failure, the error, and the repository state, the model can rewrite part of its history, replay the affected commands, and check the local change with the same test. This decomposes credit assignment into two questions that can be checked directly. Local tests determine which version to retain, and final validation assesses the result of the entire task. Local tests and text alignment identify which tokens to update and which version to prefer. The terminal training reward determines the update weights. Figure~\ref{fig:dfpo-overview} compares the sources of credit assignment in GRPO, PPO, and DFPO.

\paragraph{Our contributions.}
We propose DFPO, which combines test-validated history rewriting at inference time with token-level reinforcement learning. It generates training signals directly from a single trajectory to mitigate long-horizon credit assignment. Experiments show that the full DFPO method improves repository repair resolution rates and still outperforms GSPO under a shared history rewriting and replay protocol.

\section{Related Work}
\paragraph{Verifiable rewards and single-trajectory training.}
PPO clips the policy probability ratio to constrain updates \citep{arx1707_06347}. GRPO estimates relative advantages from multiple responses to the same problem \citep{arx2402_03300}, and GSPO uses a length-normalized sequence probability ratio \citep{arx2507_18071}. SPO initializes a reward baseline for each prompt from multiple offline samples, then generates one new trajectory whenever that prompt is selected online and updates its baseline with rewards collected for that prompt \citep{arx2509_13232}. SAO samples one trajectory per problem. In asynchronous agent training, it estimates token-level advantages with a value model and GAE that skips tool observations \citep{arx2607_07508}. DFPO constructs token-level signals from locally validated repair pairs within a single task launch and weights them by that task's terminal reward.

\paragraph{Trajectory editing and replay.}
IOP requires multiple trajectories to construct training signals \citep{arx2605_05226}. IBPO compares reasoning paths for the same problem and uses the resulting signals for sequence-level reward shaping or token-level gradient masking \citep{ding2026ibpo}. AgentRewind restores the conversation and workspace together \citep{arx2608_14380}. DFPO uses local tests to accept rewrites of the execution history, resumes the task from the revised context, and uses the same terminal reward to weight comparisons involving the original trajectory, the spliced trajectory, and repair attempts.

\paragraph{Execution feedback and process supervision.}
Code execution can provide intermediate correctness checks and failure diagnostics \citep{arx2305_18341,arx2601_20802}. ECHO uses actual terminal outputs as targets for predicting environment responses to actions \citep{arx2605_24517}. ExecCritic generates and validates tests for repository repair training \citep{arx2609_09133}. EqLen constructs equal-length reasoning segments and updates only the tokens that participate in each comparison \citep{arx2604_17328}. State commitment learning compares downstream outcomes when temporary reasoning is retained or erased, training the model to preserve the state needed for later reasoning \citep{arx2606_05201}. DFPO uses observed errors to guide local rewrites and validates revisions through replay and tests, forming contrastive supervision on assistant tokens. Appendix~\ref{app:related} covers additional related work, from policy optimization to learning from real user feedback.

\section{Direct Feedback Policy Optimization}
\label{sec:method}

\begin{figure*}[t]
  \centering
  \includegraphics[width=\linewidth]{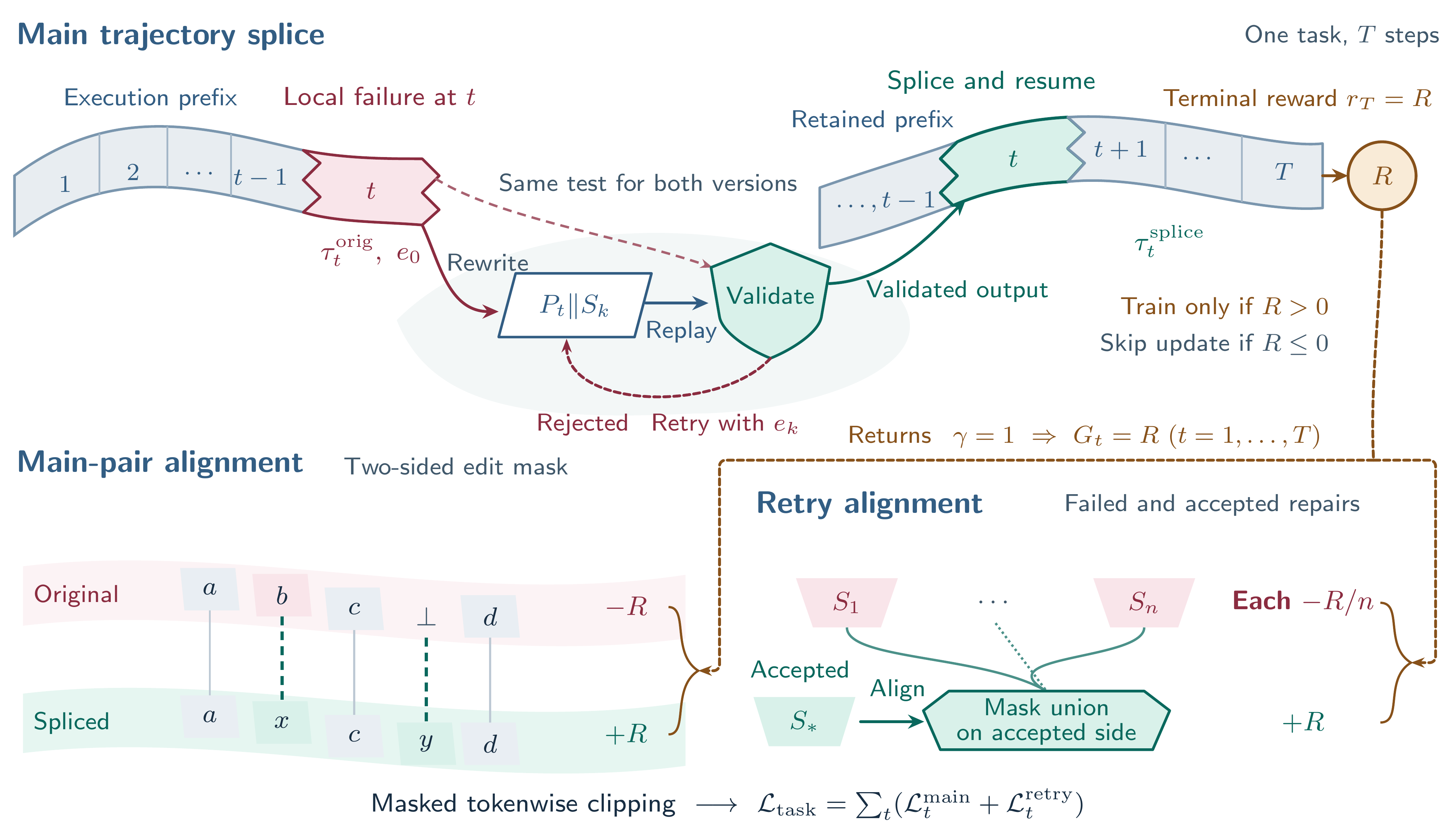}
  \caption{Trajectory splicing and token-level training in DFPO. The ribbon shows the $T$ steps of one task. The red segment marks a local test failure at step $t$. The teal segment replaces the original output at that same step. The environment gives reward $r_T=R$ only at step $T$. With discount factor $\gamma=1$, the return at every step $t$ is $G_t=R$. The repair context concatenates prefix $P_t$ and repair segment $S_k$. We apply the rewrite command in $S_k$ to the original record, replay the affected actions, and compare the original and candidate using results from the same test version. We retain the original prefix, splice in the validated assistant output (teal) with tool observations from replay, and resume execution only if the original fails, the candidate passes, and the model flags no test gaming. Dashed stitches in the lower ribbons mark substitution $b\to x$ and insertion $y$. A gap contributes no update. The retry comparison requires both a failed and an accepted version. The accepted-side mask takes the union of edited positions across all alignments. Only tasks with $R>0$ contribute updates. Gold braces show how each step’s return weights both sides of its training comparisons, and clipping applies to masked assistant tokens.}
  \label{fig:dfpo-method}
\end{figure*}

Figure~\ref{fig:dfpo-method} shows how local validation produces training pairs within a single task and how the terminal reward weights updates at the edited token positions.

\subsection{Rewriting and Repair Within a Single Execution}
In each data collection round, each coding task is launched once from the initial workspace, with one main trajectory that can be resumed after repair. Inference, testing, rewriting, command replay, and final validation share the same execution deadline. The deliverable in the repository repair experiments is a code patch.

\paragraph{Collecting failure evidence.}
At step $t$, the agent implements a feature, writes a test for it, and runs that test. If it passes, the agent moves on to the next task item. If it fails, we retain the original model output, input prefix, and actual tool observations. We denote this trajectory segment by $\tau_t^{\mathrm{orig}}$ and the test error by $e_0$.

\paragraph{Rewriting and replay.}
The repair prefix $P_t$ contains $\tau_t^{\mathrm{orig}}$, $e_0$, and other information relevant to the repair. The execution trajectory is stored in a file. Given $P_t$, the model generates a repair segment $S_1$ with its repair plan and a command $C_1$ that edits earlier assistant outputs in the trajectory file. We apply $C_1$ to a working copy of the file, replay the affected code-editing commands to apply the revision to the repository workspace, and run the local test. The local test may be revised. After a test change, we revalidate the original and revised versions with the same test version. When the test is unchanged, we reuse the original version's test result. The same policy model that generates the revision checks it for test gaming against the task requirements and the changes to the code and tests, including whether weakened assertions bypass a requirement. We accept the revision only if the original fails, the revision passes, and the check flags no test gaming. If we accept the first revision, we set $S_*=S_1$. For a rejected revision, we record the actual execution error or validation feedback as $e_1$.

\paragraph{Retrying repairs and resuming execution.}
For each retry, the model uses $e_k$ to rewrite the entire repair segment $S_k$, including its reasoning and editing command, yielding $S_{k+1}$. The repair context concatenates $P_t$ with the corresponding version. We retain the earlier failed versions and use the command in $S_{k+1}$ to construct a candidate from the original record, then replay and validate it under the same acceptance rule. We denote the accepted version by $S_*$. We splice the validated assistant output back into its original position, replace the affected tool observations with the actual replay results, and resume from this context to obtain $\tau_t^{\mathrm{splice}}$. The training pair thus shares the unchanged prefix, with each side retaining its own conditioning context and actual tool execution record.

\paragraph{Final validation.}
During training, each submission runs the training validation suite. If validation fails with time remaining, the agent continues repairing and tries again. Local validation determines whether a revision can be used to resume execution and form a training pair. We use a standard episodic reinforcement learning setup with terminal rewards, treating each coding task as an episode of $T$ steps. The scalar reward $r_k$ is zero before termination, and the terminal reward is the environment score $r_T=R$. With undiscounted returns ($\gamma=1$), the sampled return from each step $t$ is $G_t=\sum_{k=t}^{T}r_k=R$. We retain tasks with $R>0$ for updates and use each step's return to weight its local training pairs. Tasks with $R\leq 0$ are skipped during updates.

\subsection{Edit-Distance Masks}
\label{sec:mask}
DFPO uses a two-sided edit mask that marks edited positions in the original and spliced versions. Let $y_t^{\mathrm{orig}}$ and $y_t^{\mathrm{splice}}$ denote the assistant token sequences compared on the two sides of the main training pair. We compute a shortest token-level Levenshtein edit script \citep{wagner1974string}
\begin{equation}
 \begin{aligned}
  \mathcal E_t
  &=\EditOps(y_t^{\mathrm{orig}},y_t^{\mathrm{splice}})\\
  &=((u_1,v_1),\ldots,(u_M,v_M)).
 \end{aligned}
  \label{eq:edits}
\end{equation}
A substitution records a position on each side. A deletion sets $v_k=\bot$, and an insertion sets $u_k=\bot$. Matching tokens are excluded from the edit script. The positions of all edits define a mask for each side
\begin{align}
  \mathcal S_t^{\mathrm{orig}}
  &=\{u_k:1\le k\le M,\ u_k\ne\bot\},\notag\\
  \mathcal S_t^{\mathrm{splice}}
  &=\{v_k:1\le k\le M,\ v_k\ne\bot\},\notag\\
  m_{t,j}^{b}&=\mathbf 1\{j\in\mathcal S_t^b\},
  \quad b\in\{\mathrm{orig},\mathrm{splice}\}.
  \label{eq:mask}
\end{align}
The shared history and tool observations provide conditioning context for scoring. The edit masks identify the assistant tokens that contribute directly to the policy loss. Local validation determines the preference direction, and the selected tokens on each side share the contrastive weight set by the terminal score. The local test returns a pass or fail result for the candidate version. We do not claim that the edit mask identifies the causal effect of any individual edit on the test outcome.

\subsection{Paired-Trajectory Loss for the Main Task}
At the start of each collection round, we freeze the old policy $\pi_{\mathrm{old}}$. Let $a_{t,j}^{b}$ be the target token at position $j$ on side $b$ of the training pair, and let $c_{t,j}^{b}$ be its prefix on that side. The current-to-old policy probability ratio is
\begin{equation}
 \rho_{t,j}^{b}(\theta)
 =\frac{\pi_\theta(a_{t,j}^{b}\mid c_{t,j}^{b})}
        {\pi_{\mathrm{old}}(a_{t,j}^{b}\mid c_{t,j}^{b})}.
 \label{eq:ratio}
\end{equation}
For the original tokens and those sampled after the splice, we use the old policy probabilities recorded at sampling time. The training distribution of directly rewritten tokens is induced by generating rewrite commands, replaying the affected operations, and applying local validation. The model generates rewrite commands conditioned on the original output and the observed error. Once a revision is accepted, the spliced trajectory includes the rewritten assistant output and the actual tool observations from replay. At scoring time, the frozen old policy assigns the denominator of Equation~\eqref{eq:ratio}, a reference probability under the post-splice prefix $c_{t,j}^{\mathrm{splice}}$. The resulting ratio defines a clipped contrastive surrogate over the fixed repair pairs and scoring prefixes.

Let $\rho=(\rho_j)_{j\in\mathcal M}$ be the probability ratios at positions selected by mask $\mathcal M$, and let $w$ be the signed weight. For a nonempty mask, we clip the ratios at the lower bound $1-\epsilon_{\mathrm{low}}$ and the upper bound $1+\epsilon_{\mathrm{high}}$, and average the tokenwise surrogate
\begin{align*}
 \bar\rho_j &= \min\{1+\epsilon_{\mathrm{high}},\max\{1-\epsilon_{\mathrm{low}},\rho_j\}\},\\
 \Clip(\rho,w) &= \frac{1}{|\mathcal M|}\sum_{j\in\mathcal M}
                 \min\{w\rho_j,w\bar\rho_j\}.
\end{align*}
For an empty mask, $\Clip(\rho,w)=0$. Let $\rho^+$ and $\rho^-$ denote the masked ratios on the spliced and original sides, respectively. For retained tasks, local validation determines the preference direction, and the positive terminal score $R$ sets the weight magnitude. We assign $+R$ to the accepted spliced version and $-R$ to the original version, yielding
\begin{equation}
 \mathcal L_t^{\mathrm{main}}
 =-\Clip(\rho^+,+R)-\Clip(\rho^-,-R).
 \label{eq:main}
\end{equation}
Substituting $R<0$ would make the weight $+R$ on the spliced version negative and the weight $-R$ on the original version positive, reversing the preference established by local validation. We preserve this preference direction by retaining only tasks with $R>0$ for updates. For steps without a repair pair, we set $\mathcal L_t^{\mathrm{main}}=0$.

\subsection{Repair Comparisons and Total Loss}
Let $S_1,\ldots,S_n$ be the failed repair segments at step $t$, and let $S_*$ be the version that eventually passes. The passing version receives contrastive weight $+R$, and each failed version receives $-R/n$, so the failed versions' weights sum to $-R$. We align each failed version with $S_*$. Each failed version uses its own edit mask, and the passing version uses the union of the positions selected on its side across all alignments.

We score all repair segments under $P_t$. If the sampling prefix differs from the scoring prefix, we recompute the denominator with the frozen old policy under the scoring prefix to obtain a proxy ratio for clipping. With $\rho_*$ and $\rho_k$ denoting the ratios at the masked positions of the passing and failed versions, the retry loss is
\begin{equation}
 \mathcal L_t^{\mathrm{retry}}
 =-\Clip(\rho_*,+R)-\sum_{k=1}^n\Clip(\rho_k,-R/n).
 \label{eq:retry}
\end{equation}
The repair term is $0$ when the first repair passes or when no repair passes validation. Let $B$ be the number of tasks with a positive terminal score in the batch. For $B>0$, the task and batch losses are
\begin{align}
 \mathcal L_{\mathrm{task}}(\theta)
 &=\sum_t(\mathcal L_t^{\mathrm{main}}+
          \mathcal L_t^{\mathrm{retry}}),\notag\\
 \mathcal L_{\mathrm{batch}}(\theta)
 &=\frac{1}{B}\sum_{b=1}^{B}\mathcal L_{\mathrm{task}}^{(b)}(\theta).
 \label{eq:total}
\end{align}
We skip the batch update when $B=0$.

\section{Repository Repair Experiments}
\label{sec:experiments}

\paragraph{Model and training.}
We use Qwen3.5-35B-A3B as the base model and update its parameters with rank-32 LoRA \citep{arx2106_09685}, using a learning rate of $2\times10^{-5}$ for repository repair. Our main experiments use the SWE-mini harness and 400 SWE-rebench entries published before 2026 for training \citep{arx2505_20411}. Base uses the initial weights. GSPO, GSPO + BASIS, and DFPO use the same base model and harness.
We match training compute across methods by total end-to-end GPU hours.
For each trained method, we use three distinct random seeds for training and report the mean evaluation score across the three runs.
BASIS \citep{arx2605_27293} samples one rollout per selected task at each online training step and estimates the reward baseline from other tasks in the batch. Collecting the offline reference rollouts counts toward the matched GPU-hour budget.

\paragraph{Training reward.}
The training reward adds credit for passing target repair tests and subtracts a penalty for failing regression tests. Let $N_{\mathrm{F2P}}>0$ be the total number of target repair tests, $n_{\mathrm{F2P}}^{\mathrm{pass}}$ the number of those tests that pass, and $n_{\mathrm{P2P}}^{\mathrm{fail}}$ the number of regression tests that fail. Normalizing both counts by the total number of target repair tests gives the terminal reward
\begin{equation}
 R=1.5\frac{n_{\mathrm{F2P}}^{\mathrm{pass}}-n_{\mathrm{P2P}}^{\mathrm{fail}}}
 {N_{\mathrm{F2P}}}.
 \label{eq:training-reward}
\end{equation}
The penalty is zero when all regression tests pass. Passing some target repair tests can yield a positive score after the penalty, allowing tasks that fall short of strict resolution to contribute to training.

\paragraph{Evaluation.}
The SWE-rebench holdout set contains 110 problems released after March 2026. We attempt each problem three times at temperature 0.7 and report the strict resolution rate (avg@3). SWE-bench Verified contains 500 problems, with one attempt per problem \citep{sweverified}. The evaluator runs its dedicated tests only after the agent submits its final patch and withholds the results from the agent. Strict resolution requires all fail-to-pass tests to pass and all pass-to-pass tests to remain passing. We report the fraction of attempts that satisfy this criterion. The environment's scoring rule determines the terminal training reward $R$.

\begin{table}[t]
 \centering
 \caption{Strict resolution rates in SWE-mini (\%). All methods use Qwen3.5-35B-A3B.}
 \label{tab:mini}
 \begin{tabular}{lrr}
  \toprule
  Method & SWE-rebench & Verified\\
  & 110 problems & 500 problems\\
  \midrule
  Base & 17.3 & 44.2\\
  GSPO & 19.4 & 48.5\\
  GSPO + BASIS & 21.1 & 51.7\\
  DFPO & \textbf{28.5} & \textbf{62.4}\\
  \bottomrule
 \end{tabular}
\end{table}

\paragraph{Main results.}
GSPO improves on the initial base model on both benchmarks, and GSPO + BASIS improves the resolution rates further. DFPO achieves the highest strict resolution rate on both benchmarks (Table~\ref{tab:mini}).

\begin{table}[t]
 \centering
 \caption{Strict resolution rates in OpenHands SDK on SWE-bench Verified (\%).}
 \label{tab:gspo}
 \begin{tabular}{rrr}
  \toprule
  Base (Step 0) & GSPO & DFPO\\
  \midrule
  64.0 & 69.2 & \textbf{78.2}\\
  \bottomrule
 \end{tabular}
\end{table}

\paragraph{OpenHands SDK comparison.}
The OpenHands SDK experiments use Qwen3.5-35B-A3B and 2,699 OpenSWE training tasks. Evaluation uses temperature 0.7, at most 200 interaction turns per problem, and a context budget of 200K tokens. In this harness, GSPO post-training improves on the initial base model, and DFPO achieves a higher resolution rate than GSPO (Table~\ref{tab:gspo}).

\paragraph{Ablation design for history rewriting and the training objective.}
Table~\ref{tab:repair-objective-ablation} lists the groups in SWE-mini. Groups A and B use the same initial weights, test access, and task time limits. After a local test fails, both agents can use the error to generate repair actions and rerun the test. Group A edits code in the current workspace and appends the new actions to its history. Group B can edit earlier assistant outputs in its trajectory file, replay the affected commands in the workspace, and resume from the validated context. Groups C and D compare GSPO and DFPO under comparable training budgets, with the same history rewriting and replay protocol during training and evaluation.

The component ablations use full DFPO (D) as the reference. Group E retains the repair retry procedure and removes the retry loss. Group F computes the loss on all assistant tokens in the compared segments. Group G assigns a fixed positive weight to the retained tasks and keeps the contrastive form of the main and retry losses.

Masked SFT (H) reuses the repair pairs, scoring prefixes, and edit masks used for updates in group D. It uses the same base model and update budget, computing cross-entropy only on the masked tokens of the passing versions. Strict success gating (I) keeps the DFPO objective and further restricts updates to positive-score tasks that pass all target repair and regression tests.

\begin{table}[t]
 \centering
 \caption{Ablations of history rewriting, the training objective, and DFPO components in SWE-mini. Strict resolution rate is measured on the SWE-rebench holdout set (avg@3).}
 \label{tab:repair-objective-ablation}
 \setlength{\tabcolsep}{4pt}
 \begin{tabular}{@{}lp{0.34\columnwidth}cr@{}}
  \toprule
  Group & Training setting & \shortstack{History rewrite\\and replay\\at evaluation} & \shortstack{Resolution\\rate (\%)}\\
  \midrule
  A & Initial base model & Off & 17.3\\
  B & Same base as A & On & 17.6\\
  C & GSPO with rewrite & On & 21.8\\
  D & DFPO with rewrite & On & \textbf{28.5}\\
  \midrule
  E & DFPO w/o retry loss & On & 25.5\\
  F & DFPO w/o mask & On & 24.8\\
  G & DFPO with fixed $R$ & On & 26.4\\
  H & Masked SFT & On & 24.2\\
  I & DFPO, strict gate & On & 27.6\\
  \bottomrule
 \end{tabular}
\end{table}

\paragraph{Ablation results.}
Enabling history rewriting and replay at evaluation yields a small increase in strict resolution rate for the same base model (A versus B). With the evaluation protocol held fixed, GSPO post-training improves the resolution rate (B versus C). Under the shared rewriting and replay protocol during training and evaluation, DFPO outperforms GSPO (C versus D) by a larger margin than the gain from enabling rewriting and replay for the base model (Table~\ref{tab:repair-objective-ablation}).

Retaining the retry procedure and removing the retry loss (E) lowers the strict resolution rate, indicating that contrasting passing and failed repairs provides useful supervision.
The edit mask converts coarse-grained response-level signals into token-level signals, reducing the number of tokens that directly contribute to the loss. The component ablation shows that removing the edit mask lowers the strict resolution rate.
Assigning a fixed positive weight to retained tasks (G) lowers the resolution rate, suggesting that weighting tasks by their terminal scores helps in this setting.
With the same repair pairs and edit masks, the contrastive objective in full DFPO achieves a higher strict resolution rate than masked SFT (H), which learns only from tokens in passing versions.
Strict success gating (I) slightly lowers the strict resolution rate, suggesting that retaining positive-score tasks that fall short of strict resolution helps training in this setting.

\section{Discussion and Limitations}
DFPO uses test feedback to select validated revisions and the terminal reward to weight training. The token mask identifies local update positions, while the shortest edit script measures textual differences and the local tests cover only the behaviors they exercise. Replay requires reconstruction of the repository state and tool observations. Changes to external dependencies and flaky tests can undermine this consistency. For directly rewritten tokens, the statistical properties of the proxy probability ratio depend on the relationship between the rewriting process and the old policy under the scoring prefix.

DFPO counts trajectory editing, command replay, and revalidation of both versions after a test change toward the task time limit. Training also requires rescoring rewritten tokens with the frozen old policy. The A/B comparison uses a common task time limit. Actual wall-clock time depends on the amount of generation, the scope of replay, and the number of test invocations. Comparing those costs requires measured runtimes.

This study focuses on code repository repair. Our contribution is to introduce DFPO and demonstrate its effectiveness on this task. Extending DFPO to other domains remains future work. This scope limitation does not invalidate the contribution of this work.

\section{Conclusion}
DFPO uses tests to trigger and validate rewrites within a single coding task, constructing token-level supervision from the original and revised trajectories and from repair attempts. Final validation supplies the shared reward for these comparisons, and edit-distance masks focus updates on the tokens that differ. Repository repair experiments demonstrate the method's effectiveness across execution harnesses.

\section*{AI Disclosure Statement}
We used generative AI only for LaTeX typesetting and translation.

\section*{Impact Statement}
This work uses software tests to train AI coding agents and can reduce the cost of collecting separate supervision for intermediate actions. Generated code may still contain defects, and weak test suites can miss erroneous behavior outside their coverage. The method validates revisions in an isolated working copy and uses final validation to weight the training signal. Deployment still requires project-level tests and human review.

\begingroup
\setlength{\emergencystretch}{1em}

\endgroup

\clearpage
\appendix
\onecolumn
\section{Extended Related Work}
\label{app:related}

\subsection{Policy Learning and Preference Optimization}

\noindent\textbf{SFT} uses demonstration responses as targets and increases the log probability of each target token under its corresponding prefix \citep{arx2402_03300}.

\noindent\textbf{On-policy distillation / GKD} lets the student generate its own responses, then updates it using the teacher's predictions at the prefixes the student visits \citep{arx2306_13649}.

\noindent\textbf{RFT and Online RFT} filter model outputs using correctness rules and train the model to imitate responses that pass verification. The online variant continually samples from the current policy \citep{arx2402_03300}.

\noindent\textbf{REINFORCE} weights the log-probability gradients of sampled actions by sampled returns to increase the policy's expected return \citep{reinforce}.

\noindent\textbf{GAE} estimates an action's advantage by exponentially weighting subsequent temporal-difference residuals \citep{arx1506_02438}.

\noindent\textbf{PPO} constrains policy updates with a clipped current-to-old policy probability ratio and is often used with a critic and GAE \citep{arx1707_06347}.

\noindent\textbf{GRPO} samples several responses to a problem, estimates relative advantages from within-group reward statistics, and updates the model with a clipped policy objective \citep{arx2402_03300}.

\noindent\textbf{Verifiable rewards in DeepSeek-R1} compute correctness scores from reference answers or program tests and combine them with format constraints to train a reasoning policy \citep{arx2501_12948}.

\noindent\textbf{GSPO} clips a length-normalized probability ratio between the current and old policies for the full response \citep{arx2507_18071}.

\noindent\textbf{DPO} directly optimizes the generative policy with a binary preference loss based on the difference between preferred and dispreferred responses' log-probability ratios relative to a fixed reference policy \citep{arx2305_18290}.

\noindent\textbf{2-GRPO} reduces the group to two responses, applies positive and negative advantages when one succeeds and one fails, and relates this update to DPO through their contrastive gradients \citep{arx2510_00977}.

\subsection{Single-Trajectory Sampling, Baseline Estimation, and Distribution Matching}

\noindent\textbf{SPO (Single-stream Policy Optimization)} initializes each prompt's baseline with the mean reward of $n_0$ responses sampled from the initial policy before online training, with $n_0=8$ in the original experiments. At a prompt's first online visit, SPO subtracts the initial baseline from the new trajectory's reward to compute its advantage and then updates the baseline with the new reward. Later visits generate fresh trajectories, using the stored baseline and adjusting the weight of past rewards according to the policy's KL divergence \citep{arx2509_13232}.

\noindent\textbf{VAPO} improves value estimation and advantage computation for long reasoning tasks to address critic bias, variable trajectory lengths, and sparse rewards \citep{arx2504_05118}.

\noindent\textbf{SAO (Single-Rollout Asynchronous Optimization)} samples one trajectory per problem in asynchronous agent reinforcement learning, estimates token-level advantages with a value model and GAE that skips environment observations, and masks tokens whose policy deviation from the sampling policy exceeds a threshold \citep{arx2607_07508}.

\noindent\textbf{BASIS} samples one response per problem online in each round, initializes values from offline reference-policy samples, and corrects the per-problem baseline with rewards from other problems in the batch \citep{arx2605_27293}.

\noindent\textbf{SR-PPO} uses an online critic as a prefix baseline and estimates Pass@$k$ gradient weights so that training with one trajectory per problem emphasizes problems with low success rates \citep{arx2606_25451}.

\noindent\textbf{MSSR} uses a Beta baseline and adjusts advantages based on output entropy to stabilize single-trajectory training on multimodal tasks \citep{arx2512_18215}.

\noindent\textbf{Kernelized AE} applies kernel smoothing to rewards for the same problem across rounds, estimating problem value when few responses are sampled per problem \citep{arx2604_28005}.

\noindent\textbf{Shrinkage Baselines} shrink the same-problem reward mean toward the mean across problems, reducing the variance of baselines estimated from few samples of the same problem \citep{arx2511_03710}.

\noindent\textbf{ReMax} generates a sampled and a greedy response to each problem, using the greedy reward as the REINFORCE baseline for the sampled response \citep{arx2310_10505}.

\noindent\textbf{A*-PO} estimates optimal values from offline reference-policy samples, samples one online response per problem, and fits an optimal advantage with least squares \citep{arx2505_20686}.

\noindent\textbf{REINFORCE++} normalizes advantages across the training batch. The basic version supports one sample per problem, and the local-baseline variant uses multiple samples for the same problem \citep{arx2501_03262}.

\noindent\textbf{RLOO} uses the mean reward of the other responses to the same problem as a baseline. With one success and $n$ failures, the successful response has advantage $+1$ and each failed response has advantage $-1/n$ \citep{arx2402_14740}.

\noindent\textbf{OPO} uses a length-weighted mean reward across responses to the same problem as a baseline, then performs on-policy updates on samples from the current policy \citep{arx2505_23585}.

\noindent\textbf{SNIB} divides each full-response current-to-old policy probability ratio by the batch mean ratio and combines this with a batch-mean reward baseline for clipped updates \citep{snib}.

\noindent\textbf{Twisted SMC} uses a learned potential for future reward to weight and resample prefixes during decoding, approximating a specified target distribution \citep{arx2404_17546}.

\noindent\textbf{GFlowNet language-model fine-tuning} uses a subtrajectory balance objective to train the model to sample from a reward-defined target posterior while retaining multiple plausible solutions \citep{arx2310_04363}.

\noindent\textbf{BRAIn} samples responses from a proposal distribution and uses self-normalized importance sampling and a baseline to train the model to match a reward-conditioned posterior \citep{brain}.

\subsection{Trajectory Rewriting, Rollback, and Repair}

\noindent\textbf{SCoRe} learns from the model's own multi-turn correction trajectories via online reinforcement learning and uses reward improvements to encourage correction of the first response \citep{arx2409_12917}.

\noindent\textbf{IBPO (Implicit Behavior Policy Optimization)} extracts correction signals from counterfactual comparisons of reasoning paths for the same problem. It uses these signals for sequence-level reward shaping or token-level gradient masking. When the comparison signal offsets some of the terminal reward noise and the shaping weight is suitable, reward shaping can reduce the variance of centered advantages before normalization \citep{ding2026ibpo}.

\noindent\textbf{FC-SWE} resets the repository after an unsuccessful attempt, resamples conditioned on the failed patch and validation feedback, and estimates advantages using each complete attempt's own validation result \citep{arx2610_07898}.

\noindent\textbf{Counterfactual Distillation} branches at a past decision, uses reflection to generate an alternative action, and trains the model to reproduce it from the original prefix. The online version also learns the reflections that lead to success \citep{arx2603_16843}.

\noindent\textbf{AgentRewind} jointly restores the conversation context and workspace from aligned checkpoints, then resumes execution with lessons from the failed attempt \citep{arx2608_14380}.

\noindent\textbf{Generator-Assistant Stepwise Rollback} uses an auxiliary model to locate the failed step, then rolls back and replays a useful action prefix to resume execution \citep{stepwiserollback}.

\noindent\textbf{CausalFlow} replaces candidate steps in failed trajectories and replays downstream computation. It uses minimal corrections that change the final outcome to construct step-level contrastive samples \citep{arx2605_25338}.

\noindent\textbf{SWE-Replay} reuses previous exploration records at inference time and branches again from selected steps, reducing the cost of resampling full repository repair trajectories from scratch \citep{arx2601_22129}.

\noindent\textbf{Cleaning an Agent's History Hurts When the World Still Holds the Error} varies the history and environment state in controlled comparisons and finds that consistency between the rewritten history and actual environment state affects the agent's success after resuming \citep{arx2611_14654}.

\noindent\textbf{CompactionRL} jointly trains task execution and context summarization, using cross-segment advantage corrections to propagate terminal rewards to outputs before and after compaction \citep{arx2607_05378}.

\noindent\textbf{GLM-5} computes token probability ratios directly from rollout logs during asynchronous agent reinforcement learning and masks policy gradients for tokens that fall outside the allowed bounds \citep{arx2602_15763}.

\noindent\textbf{GLM-5.2} uses critic-based PPO on individual rollouts to optimize token-level advantages in long-horizon agent reinforcement learning and includes subtrajectories produced by compaction in training \citep{glm52}.

\noindent\textbf{HP-RLVR} constructs hindsight prefixes from the continuations and validation feedback of completed trajectories and weights single-trajectory updates by differences between the pre-outcome and hindsight token distributions \citep{arx2610_20497}.

\noindent\textbf{Correctable Fork Tokens} uses a branch that sees the correct answer to locate key decision tokens such as tool names and parameters, then uses verification results to assign credit to them \citep{correctablefork}.

\noindent\textbf{RLMEC} uses a minimal-edit constraint to train a generative reward model that rewrites incorrect answers and then uses that model to provide token-level reinforcement learning signals \citep{rlmec}.

\noindent\textbf{TLCR} uses Levenshtein alignment between original and rewritten responses to create token preference labels, then trains a discriminator to output continuous token rewards \citep{tlcr}.

\noindent\textbf{FIGA} compares responses of different quality and converts token- or phrase-level differences into fine-tuning supervision \citep{figa}.

\subsection{Execution Feedback and Process Rewards}

\noindent\textbf{CodeRL} trains a critic to predict the functional correctness of generated programs and uses its scores as dense feedback for program synthesis. At inference time, it uses unit test outcomes and critic scores to guide regeneration \citep{coderl}.

\noindent\textbf{RLTF} uses program-level test outcomes and error locations as rewards at different granularities in online reinforcement learning \citep{rltf}. DFPO uses local test failures to trigger history rewrites and reruns the same test to construct training pairs from failed and passing segments.

\noindent\textbf{StepCoder} breaks code generation into a curriculum of completion tasks and restricts policy updates to executed code segments \citep{stepcoder}. DFPO compares segments before and after a locally verified repair and applies edit-distance masks to the differing assistant tokens.

\noindent\textbf{Afterburner} iteratively revises programs using sandbox measurements such as runtime and peak memory. Its GRPO reward includes functional correctness and efficiency gains \citep{afterburner}. DFPO uses functional tests to verify local repairs and the final task reward to weight the resulting comparisons.

\noindent\textbf{AEGIS} turns issue descriptions into executable bug reproduction scripts \citep{aegis}. These scripts illustrate one way to obtain the local failure evidence that DFPO uses to trigger and verify trajectory rewrites.

\noindent\textbf{EqLen} constructs equal-length reasoning segments as the comparison units for group-relative reinforcement learning. A segment mask restricts updates to newly generated tokens that participate in the comparison \citep{arx2604_17328}.

\noindent\textbf{State commitment learning} compares downstream correctness with temporary reasoning retained or erased, training the model to write information needed for later reasoning into persistent state \citep{arx2606_05201}.

\noindent\textbf{CodeLift} balances test scores by grouping behaviors, compares new fixes and regressions across repository snapshots, and assigns local credit to the corresponding action segments \citep{arx2610_17224}.

\noindent\textbf{RLCSF / Lanser-CLI} converts changes in compilation, type diagnostics, and edit checks across adjacent tool actions into replayable process rewards \citep{arx2510_22907}.

\noindent\textbf{RLCF} uses a compiler to check code and a language model to compare candidate code with a reference implementation, then uses both sources of feedback to train code models with reinforcement learning \citep{arx2305_18341}.

\noindent\textbf{TaPR} runs the same task-provided hidden tests on each round, assigning process rewards based on changes in the pass rate between adjacent code versions and adding a bonus for passing all tests \citep{arx2608_00494}.

\noindent\textbf{VeRPO} weights tests by their difficulty as observed during training, using weighted partial test success in each round to provide dense rewards \citep{arx2601_03525}.

\noindent\textbf{CodeDance} directly penalizes failed Python calls at each turn and combines that signal with the answer reward for the entire trajectory \citep{arx2512_17312}.

\noindent\textbf{MIST-RL} rewards the test generator according to how many previously undetected faulty program variants a new test catches \citep{arx2603_01409}.

\noindent\textbf{ExecCritic} uses a separate model to generate tests, freezes the tests once validated, and trains the repository repair policy with feedback from repeated runs of the same test \citep{arx2609_09133}.

\noindent\textbf{TDD-Agent} first generates executable tests, then iteratively revises the implementation and tests using execution feedback \citep{arx2608_16742}.

\noindent\textbf{ZeroCoder} cross-executes self-generated programs and tests, estimates reliable candidates from the pass matrix, and jointly trains the code and test models \citep{arx2604_07864}.

\noindent\textbf{ECHO} treats actual terminal outputs as prediction targets and adds a cross-entropy loss on environment observation tokens to the policy-gradient objective \citep{arx2605_24517}.

\noindent\textbf{SDPO (execution feedback)} adds error messages and other feedback to a hindsight teacher prefix for the same model, then distills the resulting token distribution back into the original policy \citep{arx2601_20802}.

\noindent\textbf{FLARE} uses a generative reward model to diagnose executed steps, aggregating step scores into a reward for trajectory updates \citep{arx2609_23808}.

\noindent\textbf{CER} predicts final test outcomes from unfinished repository-task prefixes, tool outputs, and the current diff to score trajectories for early filtering \citep{arx2609_31995}.

\noindent\textbf{TRACE (turn-level credit)} estimates state value from changes in a frozen model's probability of the correct answer at tool-call boundaries, then constructs turn-level rewards from temporal differences \citep{arx2607_13988}.

\noindent\textbf{TIPS} uses a frozen snapshot of the current model to measure the increase in log probability of the correct answer after each retrieval and turns that increase into an information reward for the turn \citep{arx2603_22293}.

\noindent\textbf{InfoPO} compares the next-action predictions obtained with actual feedback and with the feedback masked, assigning credit to turns that yield useful information \citep{arx2603_00656}.

\noindent\textbf{ExecVerify} uses interpreter traces to verify the next statement and variable values or types, providing testable intermediate rewards for code-execution reasoning \citep{execverify}.

\noindent\textbf{MT-GRPO} converts tool-call success and retrieval quality into intermediate rewards and assigns them to the corresponding actions by turn \citep{arx2505_11821}.

\noindent\textbf{\ensuremath{\mu}Code} treats the revised program at each round as an executable candidate and alternates between training an execution-feedback-conditioned code generator and a verifier that scores candidates \citep{arx2502_20380}.

\noindent\textbf{SWE-Shepherd} constructs action-level reward data from repository-task trajectories and trains a process reward model to score candidate coding-agent actions \citep{arx2604_10493}.

\noindent\textbf{AgentPRM} uses temporal differences and GAE to construct labels and trains a process reward model to assess the likelihood of eventual task completion and the progress made by the current action \citep{arx2511_08325}.

\noindent\textbf{Are Dense Rewards Worth Their Tokens in GRPO?} compares dense and terminal rewards under matched generation-token or training-time budgets. Gains measured by update count can reverse after accounting for the cost of additional generation \citep{arx2611_13560}.

\subsection{Repository Training Data and Self-Evolution}

\noindent\textbf{SWE-bench} turns real GitHub issues and corresponding repository snapshots into repair tasks and uses project tests to check whether a patch resolves the issue \citep{arx2310_06770}.

\noindent\textbf{SWE-Gym} provides executable environments, tests, and agent trajectories for real repository tasks to train repair policies and verifiers \citep{arx2412_21139}.

\noindent\textbf{ScaleSWE} uses agents that set up environments, generate tests, and organize task descriptions to turn real GitHub pull requests into verified training tasks \citep{arx2602_09892}.

\noindent\textbf{SWE-smith} automatically synthesizes changes to real repositories that cause existing tests to fail, then uses those changes to construct verifiable repair tasks \citep{arx2504_21798}.

\noindent\textbf{From Patches to Trajectories} converts maintainer patches into solution milestones and uses them as privileged information to select useful steps in training trajectories \citep{arx2605_21996}.

\noindent\textbf{FrogNano} continually generates verifiable repository tasks near the current agent's capability limit and uses the synthetic tasks to train small coding agents \citep{arx2609_07925}.

\noindent\textbf{SWE-Growth} selects tasks that the teacher can solve with verification and on which the student still has room to improve, then uses student performance to guide subsequent task generation \citep{arx2611_23876}.

\subsection{Learning from Real User Behavior}

\noindent\textbf{DECODE} tracks the actual edits developers make after accepting AI completions. It uses the original completion and first edit to supervise prediction of the code the user ultimately keeps \citep{arx2607_25130}.

\noindent\textbf{Cursor Composer real-time RL} extracts implicit rewards from real online interactions with the current model version, updates the coding model, and deploys the new version after validation \citep{cursorcomposer}.

\noindent\textbf{GitHub Copilot Next Edit Suggestions} uses real IDE editing sessions from volunteers to supervise next-edit prediction and then uses a scorer for reinforcement learning \citep{copilotnes}.

\noindent\textbf{MAI Code 2} includes prompts and associated context from real GitHub Copilot sessions in its training data \citep{maicode2}.

\noindent\textbf{ProCodeBench} annotates developer intent from real VS Code action sequences and fine-tunes intent prediction using context such as code edits and terminal execution \citep{arx2605_05700}.

\noindent\textbf{CDHF} trains a predictor of suggestion acceptance from actual Copilot accept and reject logs to decide when to show coding suggestions \citep{arx2306_04930}.

\noindent\textbf{Code4Me and intelligent triggering} record suggestions, user choices, and the code ultimately entered. Subsequent work trains completion-trigger models on user acceptance and explicit trigger behavior \citep{arx2402_16197,arx2405_14753}.

\noindent\textbf{RECAP} combines AI conversations and fine-grained code edits from real development sessions into a replayable timeline, linking questions, adoption, and subsequent changes \citep{recap}.

\noindent\textbf{RovoDev Code Reviewer} measures adoption of a review suggestion by checking whether the subsequent commit contains the corresponding code change \citep{arx2601_01129}.

\noindent\textbf{TellOnce} compiles user corrections into scoped behavior rules and runtime checks, using check results to trigger revisions in future tasks \citep{arx2606_13174}.

\noindent\textbf{WildFeedback} identifies user dissatisfaction in real conversations and constructs preference pairs from the responses that caused it and rewrites based on the feedback \citep{arx2408_15549}.

\noindent\textbf{WildReward} converts corrections, rejections, and positive engagement in real follow-up messages into ordinal satisfaction labels, then trains a reward model and applies online DPO \citep{arx2602_08829}.

\noindent\textbf{ReSPECT} decodes positive, neutral, and negative feedback from natural human follow-ups and updates the model between deployment rounds \citep{respect}.

\noindent\textbf{Aligning Language Models from User Interactions} adds the user's next message to a hindsight teacher prefix and trains the policy using the difference between pre-outcome and hindsight log probabilities for the same original-response tokens \citep{arx2603_12273}.

\noindent\textbf{Coactive Learning} treats a user's edits to a model output as improvements and constructs preference updates from the original and edited responses \citep{coactive}.

\noindent\textbf{PRELUDE / CIPHER} infer preference instructions from historical user edits and use them to improve subsequent prompts. The experiments simulate user edits with a model \citep{arx2404_15269}.

\noindent\textbf{Principled Fine-tuning from User-Edits} uses user edits as demonstrations, relative preferences, and edit-cost rewards, combining these signals in simulated user experiments \citep{principlededits}.

\DFPOAppendixBibliography{Appendix References}
\end{document}